\documentclass{article} 
\usepackage{nips13submit_e,times}
\usepackage{amsmath}
\usepackage{hyperref}
\usepackage{url}
\usepackage{algorithmicx}
\usepackage{algorithm}
\usepackage[noend]{algpseudocode}
\usepackage{graphicx}
\usepackage{wrapfig}
\usepackage{framed}
\usepackage{siunitx}
\usepackage[sort&compress, numbers]{natbib}
\setcitestyle{numbers}

\algrenewcommand\algorithmicrequire{\textbf{Precondition:}}
\algrenewcommand\algorithmicensure{\textbf{Postcondition:}}

\usepackage{filecontents}

\title{Holophrasm: a neural Automated Theorem Prover for higher-order logic}

\author{
Daniel P.Z.~Whalen \\ 
Stanford University\\
\texttt{dwhalen@stanford.edu} \\
}

\nipsfinalcopy 

\begin{document}

\maketitle

\begin{abstract}

I propose a system for Automated Theorem Proving in higher order logic using deep learning and eschewing hand-constructed features.  Holophrasm
 exploits the formalism of the Metamath language and explores partial proof trees using a neural-network-augmented bandit algorithm and a sequence-to-sequence model for action enumeration. 
 The system proves 14\% of its test theorems from Metamath's \texttt{set.mm} module.
\end{abstract}

\section{Introduction}
Formalized mathematics arises from a desire for rigor.  A formal proof of a theorem is a proof that is complete: every step follows directly from previous steps and known theorems in an algorithmically-verifiable manner.

A number of corpora have been developed for formalized mathematics in various formalisms.  Large datasets of formal proofs include Metamath~\cite{megill1997metamath}, the Mizar Mathematical Library~\cite{matuszewski2005mizar}, Flyspeck~\cite{hales2006introduction}, the Archive of Formal Proofs~\cite{Jaskelioff-Merz-AFP05}, the Coq standard library~\cite{coq}, and the HOL Light library~\cite{harrison1996hol}.  These databases cover wide swaths of mathematics.  The Metamath \texttt{set.mm} module, for example, is a collection of theorems and proofs constructing mathematics from ZFC.  The module includes a number of important theorems, including Zorn's Lemma from set theory, the theorem of quadratic reciprocity from number theory, and Sylow's theorems from group theory.

The time cost of formalizing proofs is substantial, and so tools to assist in construction of the formal proofs have arisen.  Interactive Theorem Provers automate the technical steps of theorem-proving, leaving the creative steps to the user.  Over time, the techniques from Interactive Theorem Provers have been extended to Automated Theorem Provers, complete non-interactive tools for the generation of formal proofs.  Rapid advances are being made in Automated Theorem Proving, and 
recent systems now permit proofs of 40\% of the theorems in the Mizer Mathematical Library~\cite{kaliszyk2015mizar}.  These proof systems generally consist of multiple modules, one of which is Premise Selection: the identification of relevant axioms and theorems.  Premise Selection has shown promise as a target for machine-learning techniques~\cite{alama2014premise, kuhlwein2012overview, kuhlwein2013mash} and more recently deep learning~\cite{alemi2016deepmath} --- the first application of deep learning to Automated Theorem Proving.

While most of the current research has focused on the Mizar Mathematical Library, I demonstrate that the tree structure of Metamath proofs is exploitable by modern tree exploration techniques.
Holophrasm\footnote{Here, ``holophrasm" is the notion that a complicated idea can be conveyed by a simple theorem-vector.}
takes a novel approach to Automated Theorem Proving.  The system uses a variant of UCT~\cite{kocsis2006bandit}, an algorithm for the tree-based multi-armed bandit problem, to search the space of partial proof trees of a Metamath theorem.  Recent developments in machine learning have made such searches accessible.  Action enumeration is made viable by sequence-to-sequence models~\cite{sutskever2014sequence}. In parallel, algorithms developed for go-playing AIs describe how neural networks can be used to guide tree exploration~\cite{maddison2014move, silver2016mastering}.  Those techniques have been adapted here to create a complete, non-interactive system for proving Metamath propositions.

\section{The Metamath Format and Data Set}
The Metamath language is designed for automated theorem verification, utilizing metatheorems and proper substitutions as the standard proof step.  The exact specification of the language is given in section 4 of \cite{megill1997metamath}, but the relevant details are summarized below
\subsection{Metatheorems}

A \textit{theorem} in the Metamath database is a \textit{proposition} if it has a proof or an \textit{axiom} if it does not.  The notion of axiom here is general and includes what are traditionally known as axioms, but also includes definitions and the production rules for expressions.

We separate the axioms and propositions in \texttt{set.m} by their \textit{type}, which is ``set," ``class," ``wff," or ``$\vdash$".  Axioms of non-``$\vdash$"-type describe the production rules for a context-free grammar with the types as syntactic categories.  An \textit{expression} of the given type is a string of the corresponding syntactic category.  These axioms along with the free variables as additional terminal symbols provide a unique parse tree for every expression, and they will be referred to as \textit{constructor axioms}.  Henceforth I will conflate the notion of an expression and its parse tree.

Propositions of type ``set,'' ``class,'' and ``wff,'' are ignored to maintain uniqueness of the parse trees.

Axioms and propositions of ``$\vdash$'' type are assertions that an expression of ``wff''-type is true, that the expression can be proved from the axioms and given hypotheses.  These theorems will be used as nodes in proof trees.

\begin{wrapfigure}{R}{0.4\textwidth}
\label{proof_diagram}
  \centering
    \includegraphics[width=0.4\textwidth]{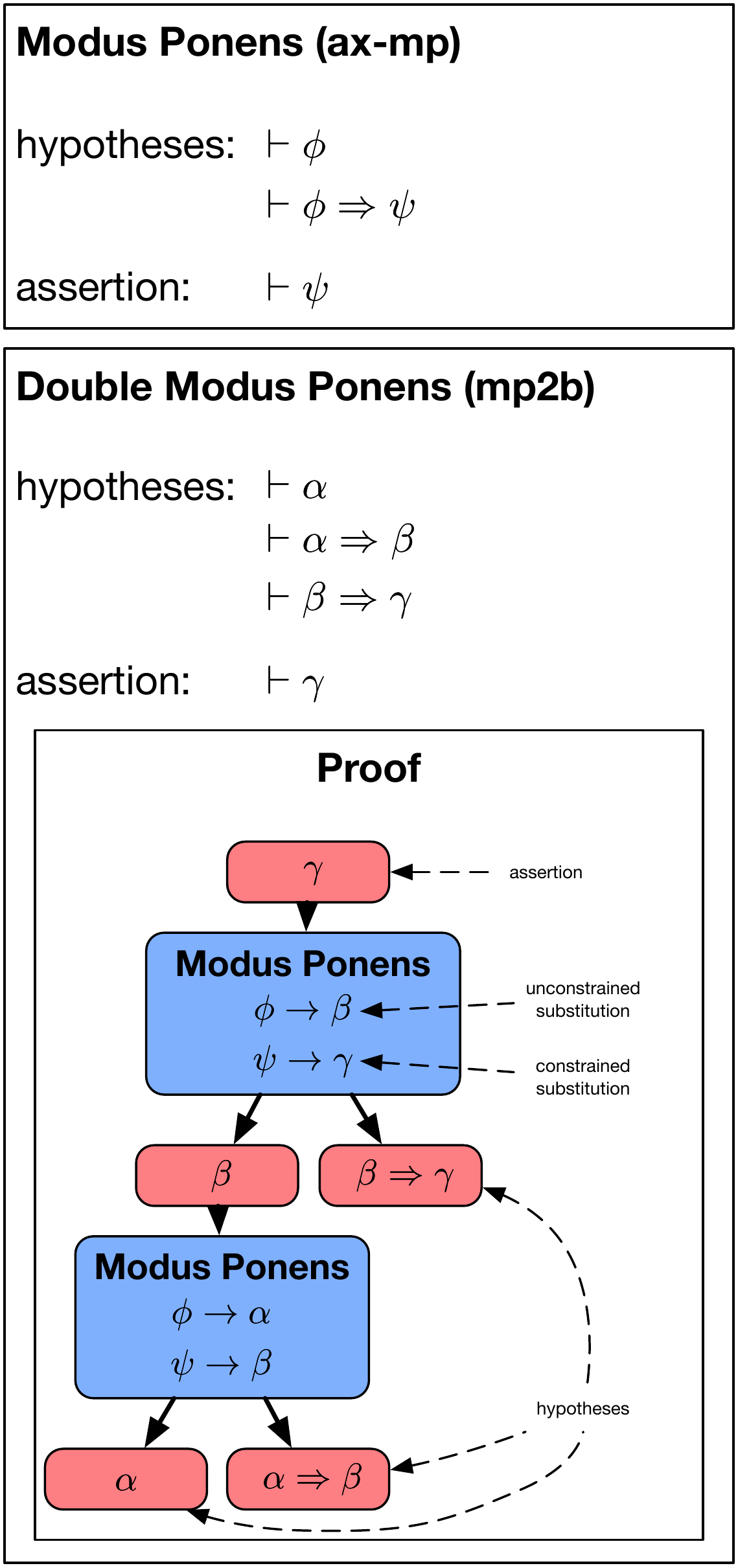}
  \caption{Examples of an axiom, {\bf {ax-mp}} and a proposition, {\bf mp2b}.  The red nodes are expressions, and the blue nodes each describe a theorem and substitutions that will prove the parent.}
\end{wrapfigure}

A theorem $T$ of $``\vdash"$-type consists of a number of elements
\begin{itemize}
\item $a_T$, an assertion, which is an expression of ``wff''-type.
\item $e_T$, a set of hypotheses, each an expression of ``wff''-type.
\item $f_T$, a set of \textit{free variables} that appear in the assertion and hypotheses and a type for each.
\item $d_T$, a set of unordered pairs of \textit{disjoint variables} from $f_T$.
\end{itemize}

The disjoint variables satisfy $(x, x)\not\in d_T$ for all $x\in f_T$, and represent pairs of variables which can not share any variables after a proper substitution.

\subsection{Proper Substitution}
Consider a context proposition $C$ that is to be proven and an expression $a$ of ``wff"-type, either the assertion of $C$ or an intermediate step in the proof of $C$.

An application of particular theorem, $T$, to prove $a$ consists of a set of substitutions, $\phi$, into the free variables $f_T$.  In these substitutions, earch variable is replaced by an expression of the same type built out of the constructor axioms extended by additional terminal symbols for the free variables of $C$.  The application requires that the assertion of $T$ after substitution matches $a$, that is $\phi(a_T)=a$.  By performing this process we reduce the problem of proving $a$ to the problem of proving all of the hypotheses $\phi(e_T)$.  This process is illustrated in figure~\ref{proof_diagram}.

The disjointness property adds a restriction on the allowable substitutions:  for every pair $(x,y)\in d_T$, for every free variable $z\in\phi(x)\cap f_C$, and every free variable $w\in\phi(y)\cap f_C$ it must be the case that $(z,w)\in d_C$.

For a fixed expression, context, and theorem, substitutions that satisfy these properties are called  \textit{viable}.  For a fixed expression and context, a theorem is called viable if it permits a viable set of substitutions.

I divide the variables in $f_T$ into two types.  \textit{Constrained} variables are variables that appear in $a_T$.   \textit{Unconstrained} variables are variables that appear in some hypothesis $h\in e_T$ but not $a_T$.  The substitutions in $\phi$ are called constrained substitutions or unconstrained substitutions if they apply to constrained or unconstrained variables respectively.  Constrained substitutions are notable in that, given $a$ and $T$, the constrained substitutions are exactly those fixed by the requirement that ${\phi(a_T) = a}$.

\subsection{Proof Trees}
\label{prooftrees}
The application of a theorem proves an expression, but also provide a set of hypotheses which must be proven in turn.  This naturally gives a proof a tree structure.  The assertion of the theorem is the root node, and its hypotheses are leaves, because they are assumed to be true without proof.  Here I define the notion of a proof tree, but I modify the natural structure slightly to permit compatibility with the notion of a partial proof tree introduced  in section~\ref{ppt}.

A \textit{proof tree} of an expression $a$ of ``wff"-type in context $C$ is a bipartite tree with two types of nodes, red nodes and blue nodes.  Red nodes are labeled by an an expression of ``wff"-type, which is an intermediate step in the proof, and the root node is labeled by $a$.  Unless its label is a hypothesis in $e_C$, in which case the node is a leaf, red nodes always have exactly one child, a blue node.  Blue nodes are labelled by a pair $(T, \phi)$ of a theorem and viable substitutions for that theorem into the parent expression.  Blue nodes have one child red node, $\phi(h)$, for each hypothesis $h\in e_T$. If such a tree exists, it is a proof of $a$.

\section{Proof Tree Exploration}\label{proofsearch}

The problem I wish to solve is as follows: given a context theorem $C$, find a proof tree for that theorem's assertion.  The algorithm does so by considering a supertree of potential proofs steps and by using tree exploration techniques to search for the subtree that is a valid proof-tree.  The algorithm will refer to three neural networks, {\bf payoff}, {\bf generative}, and {\bf predictive} which are described in section~\ref{networks}.

\subsection{Partial proof trees}\label{ppt}
A \textit{partial proof tree} is an extension of the the notion of a proof tree 
The following changes are made:  Red nodes are permitted to have no children even if they are not hypotheses.  Red nodes are permitted to have multiple multiple child blue nodes, each a potential approach for proving the expression.  

A red node is said to be proven if any of its children have been proven or if its label is one of the initial hypotheses.  A blue node is said to the proven if all of its children have been proven.

The subtree of a proven red node is necessarily a supertree of a valid proof tree for its expression.  In particular, the subtree can be pruned by removing all of the children of red nodes except for one proven blue child.  Such a pruned subtree must be valid proof tree.

\subsection{Exploration}
Similarly to UCT~\cite{kocsis2006bandit}, the algorithm builds a partial proof tree over a series of passes.  Each pass traverses the tree downward.  At a red node, the traversal chooses either to create a new child or to proceed to the highest valuation child blue node.  At a blue node, the traversal proceeds to the worst-performing child.  The pass continues until a new child blue node is created, whereupon its red children are created and valued.  The process repeats until the root node has been proven.

In order to to perform this exploration, each node maintains additional state, which is updated whenever the node's children are updated.

Red nodes, $a$ have an \textit{initial payoff}, $y_a$ which is the output of the {\bf payoff} network, evaluated as soon as the node is created.  They additionally have a \textit{total payoff}, $x_a$, which is the the sum of the initial payoff of the node and the total payoffs of its children, a \textit{visit count}, $n_a$, which is 1 plus the sum of the visit counts of its children, and an \textit{average payoff}, which is $x_a/n_a$.

Blue nodes keep track of their \textit{least promising child}, which is the unproven child with the lowest average payoff.  The total payoff and visit count of a blue node are the corresponding values of its least promising child.  Blue nodes also have a value $v_b$, which indicates how likely this substitution is to be applicable and is given by the {\bf relevance} and {\bf generative} networks.

\subsection{Visiting Nodes}

In standard UCT, when the traversal reaches a leaf, the leaf spawns a new child for each available action at that node, but doing so is impractical in this context.  In this variant, the number of actions available at red nodes can be infinite, since there are infinitely many unconstrained substitutions that could be made into some theorems.  The difficult part of the calculation is determining viable actions rather than calculating payoff.  To this end, we attempt to maintain the number of children of a red node, 
$a$ to $\lceil\frac{n_a}3\rceil$, so that more actions are considered after consecutive visits.

When a red node is visited for the first time, the node calculates its initial payoff but does not generate any children. When the node is visited later, it checks
if it has sufficiently many children.  If so the algorithm visits an extant chid as described in section~\ref{redvisit}.  If not, the algorithm creates a new child as described in section~\ref{expand}.

When a blue node is visited, it immediately visits its least promising child.

\subsection{Visiting current children of red nodes}\label{redvisit}
To determine which child is visited from a red node, each extant child $b$ is assigned a priority,
\[\frac{x_b}{n_b} + \beta\frac{v_b}{n_b}+\alpha \sqrt{\frac{\log n_a}{n_b}},\]
for constants $\beta$ and $\alpha$.  The highest priority child is then visited.

The $\frac{x_b}{n_b} + \alpha\sqrt{\frac{\log n_a}{n_b}}$ arises in the standard UCB algorithm as the upper confidence bound and the correction $\frac{v_b}{n_b}$ encourages exploring propositions with a high probability first~\cite{silver2016mastering}.  During the experiment, $\alpha$ and $\beta$ were assigned the values 1.0 and 0.5 respectively.

\subsection{Expansion of red nodes}\label{expand}
The children of a red node are constructed by assigning to them a theorem and substitutions.  Each pair $b = (T, \phi)$ of theorem $T$ and substitution $\phi$  is assigned a value $v_b =\frac{p_T} {p_\text{best theorem}}   \frac{p_{T, \phi} }{p_{\text{T, best substitution}}} $.  Here $p_T$ is the probability that $T$ is the next theorem to apply, as given by the {\bf relevance} network.  $p_\text{best theorem}$ is correspondingly the probability of the best theorem.  For a fixed $T$, $p_{T, \phi}$ is the probability of those substitutions as given by the {\bf generative} network.  $p_{T, \text{best substitution}}$ is correspondingly the probability of the best substitution.  Children of the red node are added from this expansion queue in decreasing order of value.

Evaluation of the the {\bf relevance} and {\bf generative} network are performed in a just-in-time manner, evaluating {\bf relevance} during the second visit to a node and evaluating {\bf generative} only when a previously unconsidered theorem is due to be added as a child.

When a blue node is added to the child in this way, the algorithm immediately visits each of the blue node's children once to estimate their payoffs.

\subsection{Other details} 
In practice a few additional changes can be made to make the algorithm more efficient.

{\bf Circularity}:  Any attempt to create a red node with the same expression as one its ancestors fails: the  parent blue node is removed from the expansion queue of its parent and the next pair $(T,\phi)$ is added instead.


{\bf Node Death}: While the theoretical number of actions from a given red node is infinite, in practice the number of actions is limited by the beam search width of the {\bf generative} network.  This may lead to instances where a red node has no children and its expansion queue is empty.  Such a red node is called \textit{dead}.  A blue node is said to be dead if any of its children are dead.  Dead blue nodes are removed from the graph and their ancestors are subsequently checked for death.

{\bf Multiprocessing}: The algorithm can be run efficiently in parallel.  When doing so, the different threads traverse the proof tree asyncronously.  Following \cite{silver2016mastering}, the priority function for  a blue node $b$ from a red node is modified by replacing the $\frac{x_b}{n_b}$ term with $\frac{x_b}{n_b +\gamma t_b}$, where $t_b$ is the number of threads currently exploring a descendent of $b$ and $\gamma$ is a constant, chosen here to be 3.

{\bf Generative length limits}: When evaluating the {\bf generative} network, outputs with more than 75 tokens in total across all unconstrained substitutions are discarded during the beam search.  The beam search returns no substitutions if all items in the beam reached the size limit.  If so, a dummy child is added to the red node with a payoff of 0 to discourage further exploration of this node 

{\bf Last step}: When a red node is added when proving the context $C$, the viable theorems are determined.  For the viable theorems, $T$, if there are viable substitutions $\phi$ such that $\phi(e_T)\subset e_C$ then that theorem and those substitutions are immediately added as a blue node.


\section{Networks}\label{networks}

Three distinct neural networks were used in the algorithm.  The {\bf payoff} network estimates the payoff of red nodes.  The {\bf relevance} network predicts which theorems will be useful at a given step.  The {\bf generative} network generated unconstrained substitutions.

\subsection{Tokenization}
A token is created for each constructor axiom.  A number of dummy variables are created and assigned tokens, the minimum number such that for each theorem the numbers of ``set", ``wff", ``class" free variables are at most the number of dummy variables of the corresponding type.  Five special tokens are added, `EOH' for the end of a hypothesis, `EOS' for the end of a section, `START' for the start of sequence generation, `UV' for an unconstrained variable, and `TARGET' for a target unconstrained variable. 

The inputs are modified for each iteration by randomly replacing the free variables that appear with distinct dummy variables of the corresponding type.  Each expression is tokenized by reading the tokens of the constructor axioms in its parse tree in a depth-first pre-order.  Hypotheses are separated by the `EOH' token.  If multiple different components are inputted, such as an assertion and set of hypotheses, they are separated by the `EOS' token.  The other three special tokens are used only by the {\bf generative} network and are are described later.

\subsection{Neural Networks}

The networks all share a number of similar features.  In general, the embedding vectors for tokens were inputted into 2 layers of bidirectional GRUs with internal biases for the GRU units and hidden layer dimensions of size 128.  GRU weights were permitted to vary between different sections of input and output, but the token embedding vectors were shared.  The embedding vectors were augmented with four additional dimensions describing the graph structure of the input, the depth of the node, the degree of the node, the degree of its parent, and its position into the degree of its parent. All fully-connected layers used leaky RELUs with $\alpha=0.01$ and had dimension 128 unless otherwise specified.  Weights were regularized by their $L2$-norm with a regularization factor of $\num{e-4}$.

\subsection{Payoff Network}
The {\bf payoff} network takes as input an expression, $a$, and a set of hypotheses $e_C$, which are fed into the GRUs.  The network attempts to predict whether the expression is provable from the hypotheses.  The outputs of both sides of the bidirectional network are concatenated and fed through two fully connected layers with leaky RELU units and a fully connected layer with a sigmoid to obtain the classification probability, $p_x$.

The network is trained on known proof steps as positive examples and on incorrect proof steps generated by the {\bf relevance} and {\bf generative} networks as negative examples.  During training, the cross-entropy loss is minimized.

\subsection{Relevance Network}
The {\bf relevance} network takes as input an expression, $a$, and a set of hypotheses $e_C$, and attempts to classify the next proposition that will used in the proof of $a$.  The {\bf relevance} network is designed as two parallel networks.  The first parallel branch takes $a$ and $e_C$ as inputs and returns a 128-dimensional expression-vector $v$.  The second parallel branch is evaluated separately for all theorem $T$ of ``$\vdash$"-type, inputs $a_T$ and $e_T$, and returns a 128-dimensional theorem-vector $w_T$.  The probabilities are computed as the softmax over theorems $T$, $p_T = \operatorname{softmax}(l_T)$, where $l_T = v^TWw_T$ for some weight matrix $W$.  
This structure permits generalizability to new theorems while simultaneously allowing for the theorem vectors to be precomputed and cached.

The network is trained using a negative-sampling loss with four negative samples; at each iteration, only five theorems are considered: the correct theorem $T_C$ and four incorrect theorems $T_{W,i}$ chosen uniformly at random from the viable theorems for $a$.  The training loss is computed as $-\log\sigma(l_{T_C}) - \sum_i\log\sigma(-l_{T_{W,i}})$ and is minimized.

\subsection{Generative Network}
Given an expression, $a$, a set of hypotheses, $e_C$, and a theorem, $T$, to be applied, the {\bf generative} network uses a sequence-to-sequence model \cite{sutskever2014sequence} with an intermediate fully-connected layer to create expressions for the unconstrained substitutions.

To execute the network, an unconstrained variable in $f_T$ is chosen uniformly at random to be the target.  A set of substitutions $\phi$ is generated as follows.  For $v\in f_T$ a constrained variable, $\phi(v)$ is the expression needed for $\phi(a_T)=a$.  $\phi$ additionally maps the target unconstrained variable to the `TARGET' special token and the other unconstrained variables to the `UV' special token.  The sequence-to-sequence model is used to generate an expression for $\phi$ applied to the target variable.  $\phi$ is updated to include this as a substitution, a new target unconstrained variable is chosen, and the process repeats until all variables have substitutions.

 The network takes as inputs $\phi(e_T)$ and $e_C$.
A fully-connected layer is applied to the outputs of each direction and the result is used as the initial state of the GRUs for the sequence-to-sequence output.  An attentional model is added, following the \textit{general} model of \cite{luong2015effective}.  The output sequence is initialized with the `START' token.  The outputs of the last GRU layer are fed through a fully-connected layer with RELU nonlinearity and then a fully-connected layer with softmax nonlinearly to obtain token probabilities.  During training, the total cross-entropy loss of the output tokens is minimized.
 
During execution,  multiple outputs are given, following the beam search technique of \cite{sutskever2014sequence}.  The tokens which can be included are restricted by a number of filters.  Only constructor axioms defined before the current context may be used.  No ``wff" or ``class'' variables may  be added unless they appear elsewhere in the context.  At most one new such set variable is considered during selection for a given token.  Furthermore, no token may be added if doing so would violate the disjoint variable conditions.

\section{Experiment}

\subsection{Dataset}

The theorems of the Metamath \texttt{set.mm} module are used as the data set, discarding axioms and keeping only propositions of ``$\vdash$"-type.  Of these propositions, 21786 were selected as a training set, 2711 as a validation set and 2720 as a test set.  The proofs are expanded into a full proof tree, and each step of ``$\vdash$"-type was recorded.

The {\bf relevance} network was trained and evaluated on all of the proof steps for the propositions.  The expansion of the propositions into proof steps provides 1.2M
training proof steps, 120k validation proof steps and 158k testing proof steps.

The {\bf generative} network was trained on the proof steps where a proposition was applied that had at least one unconstrained variable.  This constraint leaves 426k training proof steps, 38k validation proof steps, and 56k testing proof steps.

Data for the {\bf payoff} network was generated by including all of the proof steps as positive examples excluding duplicates.  Additionally, negative examples were generated by using the trained {\bf relevance} and {\bf generative} networks to predict the best two proposition/substitution pairs using the valuation described in section~\ref{expand}.  The hypotheses generated by applying these propositions were included as negative examples after removing all the the hypotheses that were equivalent to positive examples. There were 587k positive and 960k negative training examples, 69k positive and 113k negative validation examples, and 74k positive and 120k negative test examples.

\subsection{Training}
Network weights were initialized with Xavier initialization \cite{glorot2010understanding}.  Training for each network was done using stochastic gradient descent with a batch size of 100, with an Adam optimizer \cite{kingma2014adam}.  Learning rates started at $\num{e-4}$, were decayed by a factor of 2 each time the validation loss failed to decrease, and training was ended after the validation loss failed to decrease for three consecutive epochs.

\subsection{Performance}

The neural networks are trained separately.    The {\bf relevance} network is tested, selecting from all viable propositions.  On the test data {\bf relevance} obtains a 55.3\% top-1 accuracy, a 72.8\% top-5 accuracy and an 87.4\% top-20 accuracy.

The {\bf generative} network has a perplexity of 2.08 on the test set, when selecting from 1083 tokens.  I also measure the probability that a beam search creates the correct substitutions for all unconstrained variables as one of the results.  On the test set, {\bf generative} achieves an accuracy of 39.1\% with a beam width of 1,  51.3\% with a beam width of 5, and 57.5\% with a beam width of 20.

The {\bf payoff} network achieves a classification accuracy of 77.6\% on the test set.  For comparison, a baseline prediction of negative achieves a 62.1\% accuracy.  

The system as a whole was tested on each test theorem by expanding the proof trees for 10000 passes or until 5 minutes had passed.  Multiple attempts were made for each proposition with the beam search width set to 1, 5, or 20.    Under these parameters, the system finds proofs for 388 of 2720, or 14.3\% of the test propositions.  The system works particularly well on the initial part of the database, finding proofs for 45.1\% of the 457 test propositions in the first 5000 theorems.

In the cases that a valid proof is generated, the system works quickly.  The discovered proofs were created with a median of 17 passes.  

\section{Discussion}

In this paper I have proposed a nonconventional approach to Automated Theorem Proving for higher-order logic, and tested performance on the Metamath \texttt{set.mm} module.  While the system does not achieve state-of-the-art performance, it is the first effective complete Automated Theorem Prover to not exploit hand-crafted features.

Holophrasm takes a unconventional approach to automated theorem proving, attempting to emulate the processes and intuition of human proof exploration.  A number of new techniques and novel adaptions of current technologies have been introduced:
\begin{itemize}
\item tree-based bandit algorithms for proof exploration
\item tree-reduction during exploration passes to permit actions to have multiple subtrees
\item deep networks for estimating statement provability
\item the theorem-vector encoding for rapid theorem selection
\item sequence-to-sequence models for enumeration from an infinite set of actions
\end{itemize}
While the results of Holophrasm are not directly comparable to current results on the Mizar dataset, the developments show promise as generalizable techniques.  They highlight the feasibility of deep learning as an approach to Automated Theorem Proving.

\subsubsection*{Acknowledgments}
I am grateful to Yu-hsin Chen, Zach DeVito, and Matthew Fisher for invaluable discussions during the planning stages of this project.

\subsubsection*{References}

\small{
\renewcommand{\section}[2]{}%

\bibliography{jobname}

}

\end{document}